\documentclass{article}
\usepackage[OT1]{fontenc}
\usepackage{carr_preprint,times}

\usepackage{amsmath,amsfonts,bm}

\def\eqref#1{equation~\ref{#1}}

\def\1{\bm{1}}

\DeclareMathAlphabet{\mathsfit}{\encodingdefault}{\sfdefault}{m}{sl}
\SetMathAlphabet{\mathsfit}{bold}{\encodingdefault}{\sfdefault}{bx}{n}

\usepackage{amsmath,amssymb}
\usepackage{booktabs}
\usepackage{multirow}
\usepackage{xspace}
\usepackage[colorlinks=true,linkcolor=black,citecolor=blue,urlcolor=blue]{hyperref}
\usepackage{url}
\usepackage{graphicx}
\usepackage{float}
\usepackage{algorithm}
\usepackage{algpseudocode}

\PassOptionsToPackage{table}{xcolor}
\usepackage{xcolor,colortbl}
\definecolor{carrred}{HTML}{C0392B}
\definecolor{carrorange}{HTML}{D68910}
\definecolor{carrblue}{HTML}{2E6DA4}
\definecolor{carrpurple}{HTML}{6C4A9E}
\newcommand{\method}{\textnormal{CaRR}\xspace}
\newcommand{\tokenguard}{\textsc{TCR}\xspace}
\newcommand{\trajsubmit}{\textsc{SCR}\xspace}

\definecolor{carrgreen}{HTML}{568132}

\title{Closing Performance Gap for Diffusion Language Models under Low Inference Budgets}

\author{%
  Hao Wu\textsuperscript{1}, Sipeng Chen\textsuperscript{2}, Jingwen Fu\textsuperscript{3,4,$\dagger$},
  Yuwang Wang\textsuperscript{5,$\dagger$}\\[0.4em]
  {\normalfont\small \textsuperscript{1}\,University of Wisconsin--Madison
  \qquad \textsuperscript{2}\,Northwestern University}\\
  {\normalfont\small \textsuperscript{3}\,Zhongguancun Academy
  \qquad \textsuperscript{4}\,Zhongguancun Institute of Artificial Intelligence}\\
  {\normalfont\small \textsuperscript{5}\,Tsinghua University}%
}
\hypersetup{pdftitle={Closing Performance Gap for Diffusion Language Models under Low Inference Budgets},
  pdfauthor={Hao Wu, Sipeng Chen, Qiancheng Yang, Jingwen Fu, Yuwang Wang}}

\begin{document}

\maketitle
\begingroup
\renewcommand\thefootnote{}
\footnote{\textsuperscript{$\dagger$}Corresponding authors: \texttt{fujingwen@bza.edu.cn}, \texttt{wang-yuwang@tsinghua.edu.cn}.}
\addtocounter{footnote}{-1}
\endgroup

\maketitle

\begin{abstract}
Diffusion language models offer an architectural route to low-budget generation by updating multiple tokens in parallel, which can reduce the number of sequential decoding steps relative to token-by-token autoregressive decoding. However, realizing this advantage at low budgets is challenging, as generation quality often declines when the number of denoising and refinement steps is reduced. Editable diffusion models address this quality loss through post-draft refinement, but useful intermediate results may still be lost before final submission. We introduce \method{} (Commitment-aware Refinement and Recovery), a training-free framework which improves submission quality from a fixed editable checkpoint through updated-context refill and trajectory-based recovery within its existing forward schedule. Token Commitment Remasking (\tokenguard) remasks selected native pure replacements and refills them on the next scheduled edit forward using the updated context. Submission Commitment Recovery (\trajsubmit) uses the resulting refinement trajectory to recover supported candidates when the final response does not satisfy task-specific output requirements. Both modules operate within the native mask/edit schedule without additional backbone forwards. Experiments on code generation and mathematical reasoning show consistent gains in average accuracy over native ME-DLM across four inference budgets, with the largest gains under tight budgets. At one-eighth of the full scheduled forward budget, \method{} improves the six-metric average accuracy by 7.44 pp, including gains of 11.37 and 10.05 on MBPP and MBPP+.
\end{abstract}

\section{Introduction}
\label{sec:introduction}

Diffusion language models (DLMs) update multiple tokens in parallel, offering an architectural route to low-budget generation relative to token-by-token autoregressive decoding~\citep{nie2025llada,gong2025diffullama,ye2025dream}. However, parallel updates alone do not guarantee lower inference cost or latency. Studies of standard DLM decoding report that repeated full-sequence processing can make inference slower than autoregressive decoding at comparable output quality~\citep{wu2026fastdllm,israel2025apd,hu2026flashdlm}. Reducing the number of denoising steps lowers sampling cost, but can also degrade task accuracy~\citep{wu2026fastdllm,hu2026flashdlm}. For DLMs, this motivates improving task accuracy under a limited budget of model forwards. We study a chosen editable checkpoint and its existing refinement schedule: how much quality can be recovered while preserving its parameters and scheduled forward count?

Iterative refinement offers a way to recover generation quality by revisiting uncertain predictions. Remasking methods revisit uncertain token decisions in later steps; most closely related, T2M delays commitment by remasking low-confidence committed tokens for subsequent reconstruction~\citep{yao2026t2m}. Edit-based refinement instead revises a complete draft, as in ME-DLM~\citep{ren2026medlm}, which applies parallel replacement, insertion, and deletion operations to the current response. Yet useful information can still be underused: because parallel edits are predicted from the same pre-edit response, an immediate replacement cannot exploit simultaneous changes to its surrounding context, while terminal-only submission can discard a usable candidate that appeared earlier in the refinement trajectory. This leaves a common challenge: how to make better use of the existing refinement trajectory to improve both local editing decisions and final output selection under a fixed forward budget.

Our insight is to view editable refinement as involving two connected commitment decisions: when to commit a local token replacement during refinement, and which response to submit after refinement. Recognizing that a current token is unreliable does not imply that its immediate replacement is reliable, and completing refinement does not make the terminal response the only useful source of submission evidence. To improve accuracy under limited inference budgets, we introduce \method{} (Commitment-aware Refinement and Recovery), a training-free framework that couples token remasking with trajectory-based output recovery. The two stages are sequentially connected: token-level interventions change the response trajectory that submission-time recovery subsequently examines.

Token Commitment Remasking (\tokenguard) uses the model's support for the existing token to select native pure replacements for deferred filling. Instead of immediately committing the replacement tokens, it remasks the selected positions and refills them on the next scheduled edit forward, allowing the predictions to incorporate changes to the surrounding context. This adapts the detect--then--remask principle of T2M~\citep{yao2026t2m} to editable refinement. Submission Commitment Recovery (\trajsubmit) uses the resulting refinement trajectory to recover supported candidates when the final response does not satisfy task-specific output requirements. It follows a final-first policy, using task-specific output contracts to determine recovery from the terminal response or earlier states. Code recovery uses AST parsing and public-interface constraints, while Math recovery uses answer normalization, equivalence, and trajectory consensus. Both modules operate within the native mask/edit schedule without additional backbone forwards.

Experiments on code generation and mathematical reasoning show that \method{} improves average accuracy over native ME-DLM across four inference budgets, with the largest gains under tight budgets. In particular, at one-eighth of the full scheduled forward budget, \method{} improves the average accuracy across six metrics by 7.44, including gains of 11.37 and 10.05 on MBPP and MBPP+. Component ablations show that both modules are effective individually and provide stronger gains when combined. Mechanism analyses further support the roles of updated-context remasking and trajectory-based recovery. Our contributions are summarized as follows:
\begin{itemize}
    \item We formulate low-budget editable refinement as two connected commitment decisions: when to commit a local replacement and which response to submit after refinement.
    
    \item We propose \method{}, which improves submission quality for a fixed editable checkpoint by coupling updated-context refill of selected native replacements with final-first trajectory recovery within the native forward schedule.

    \item We demonstrate higher average accuracy over native ME-DLM across
    four inference budgets on code generation and mathematical reasoning,
    with the largest gains under tight budgets. Component and mechanism
    analyses support the complementary contributions of the two modules.
\end{itemize}

\section{Related Work}
\label{sec:related-work}

\paragraph{Diffusion Language Models and Editable Refinement.} Discrete diffusion models provide a non-autoregressive approach to text generation through corruption and iterative denoising~\citep{austin2021d3pm,lou2024sedd,sahoo2024mdlm}, and subsequent work has scaled this paradigm by adapting autoregressive checkpoints or training large masked diffusion models directly~\citep{gong2025diffullama,nie2025llada,ye2025dream}. Iterative non-autoregressive refinement also predates modern DLMs. Mask-Predict repeatedly regenerates low-confidence positions, while Levenshtein Transformer and EDITOR refine hypotheses through explicit edit operations~\citep{ghazvininejad2019maskpredict,gu2019levenshtein,xu2021editor}. Recent DLMs further increase revisability by preserving intermediate predictive information. Soft-Masked DLMs retain information from previous predictions, while EvoToken-DLM replaces hard mask-to-token transitions with progressively evolving token representations ~\citep{hersche2025softmask,zhong2026evotoken}. ME-DLM instead generates a complete draft and refines it through replacement, insertion, and deletion operations~\citep{ren2026medlm}. Our work adopts this editable post-draft setting and studies how token-level refinement and the resulting response trajectory can be better used under a fixed inference budget.

\paragraph{Low-Budget Diffusion Inference.} The efficiency of DLMs depends on maintaining generation quality under tight inference budgets. Theoretical analysis shows that the number of sampling steps required by masked diffusion can depend strongly on the evaluation objective, with sequence-level correctness exhibiting a less favorable efficiency trade-off than perplexity~\citep{feng2025theory}. Empirically, Fast-dLLM reports quality degradation under aggressive parallel decoding due to disrupted token dependencies~\citep{wu2026fastdllm}, while FlashDLM shows that reducing the number of denoising steps can degrade generation quality under standard sampling strategies~\citep{hu2026flashdlm}. Adaptive Parallel Decoding, entropy-bounded unmasking, and KLASS improve the quality--efficiency trade-off by adapting token updates across denoising steps~\citep{israel2025apd,benhamu2025eb,kim2025klass}, while dLLM-Cache reduces redundant computation through adaptive caching~\citep{liu2025dllmcache}. Similar quality--budget trade-offs have also motivated efficient sampling and few-step generation in image diffusion~\citep{song2021ddim,lu2022dpmsolver,
salimans2022progressive,song2023consistency}. CaRR complements these efficiency-oriented approaches by improving the reliability of post-draft refinement and submission under a fixed native inference budget.

\paragraph{Remasking and Commitment Control.} Another line of work restores revisability after token predictions begin to solidify. ReMDM allows generated tokens to be remasked and updated again~\citep{wang2025remdm}, while RemeDi learns confidence signals for identifying low-quality tokens and resampling them in later steps~\citep{huang2026remedi}. CoRe detects context-brittle predictions through masked-context perturbations~\citep{zhai2026core}. Most closely related, T2M identifies low-confidence committed tokens using the model's probability of the current token, remasks them, and reconstructs them in later denoising steps; controlled comparisons further separate the token detector from the remasking action~\citep{yao2026t2m}. Related approaches defer uncertain commitments, learn commitment policies from decoding traces, or coordinate multiple token commitments~\citep{shu2026dcd,sun2026tracelock,yao2026cocommit}. RDD further introduces backtracking and selective remasking to revise earlier block-level commitments~\citep{wang2026rdd}, while recent analysis shows that commitment order can directly affect reasoning quality~\citep{yeom2026answerfirst}. These studies establish the importance of revisability during diffusion generation. \tokenguard adopts this commitment-deferral principle in a different decoding regime: it intervenes on pure replacements selected by the native ME-DLM edit decoder, preserves native INSERT/DELETE actions, and refills selected positions on the next already-scheduled edit forward without increasing the scheduled NFE. The resulting editable-refinement trajectory is subsequently available to \trajsubmit for submission-time recovery.

\paragraph{Trajectory-Aware Diffusion Reasoning and Recovery.} Recent studies increasingly treat intermediate diffusion states as useful signals beyond their role in producing a terminal sample. DCoLT treats reverse-diffusion states as latent reasoning actions and optimizes the full trajectory with outcome-based reinforcement learning~\citep{huang2025dcolt}. Denoising process rewards provide process-level supervision based on the contribution of intermediate denoising intervals~\citep{xie2026dpr}, while SDPO decomposes trajectory alignment into stepwise preference objectives ~\citep{han2026sdpo}. TIE tracks confidence dynamics across intermediate states and transfers promising partially denoised trajectories between models ~\citep{yun2026tie}, while FReDA uses model-generated drafts as implicit intermediate states for recursive refinement~\citep{sun2026freda}. Prophet provides complementary evidence that useful answer information can emerge before terminal decoding and exploits this property for early commitment ~\citep{li2026prophet}. Together, these studies show that intermediate states can retain useful information before the terminal output is formed. \trajsubmit uses the completed refinement trajectory as submission-time evidence, recovering supported candidates when the native final does not satisfy task-specific output requirements.

\section{Method: \method{}}
\label{sec:method}

\subsection{Problem Definition}
\label{sec:problem}

We study how to improve the final submission of a frozen editable diffusion language model under a fixed forward budget. Native ME-DLM first generates a complete draft and then revises it through parallel replacement, insertion, and deletion operations~\citep{ren2026medlm}. The editing process can improve the draft, but a useful answer need not remain available in the terminal response.

Following ME-DLM's edit-iteration notation, let $x^{(0)}$ be the initial draft and $x^{(t)}$ the response state after $t$ edit updates. With $T$ scheduled edit forwards, the response evolves along
\begin{equation}
    x^{(0)} \longrightarrow x^{(1)} \longrightarrow \cdots
    \longrightarrow x^{(T)}.
    \label{eq:native-trajectory}
\end{equation}
Native decoding extracts its submission from $x^{(T)}$. Our objective is to improve that submission by using both the editing process and the completed trajectory. We write the input prompt as $q$; throughout, $t$ denotes an edit iteration rather than a diffusion noise level. The initial mask-diffusion phase uses $M$ forwards, giving a total scheduled budget of $M+T$.

\subsection{Framework Overview}
\label{sec:framework}

\begin{figure*}[t]
    \centering
    \includegraphics[width=\textwidth]{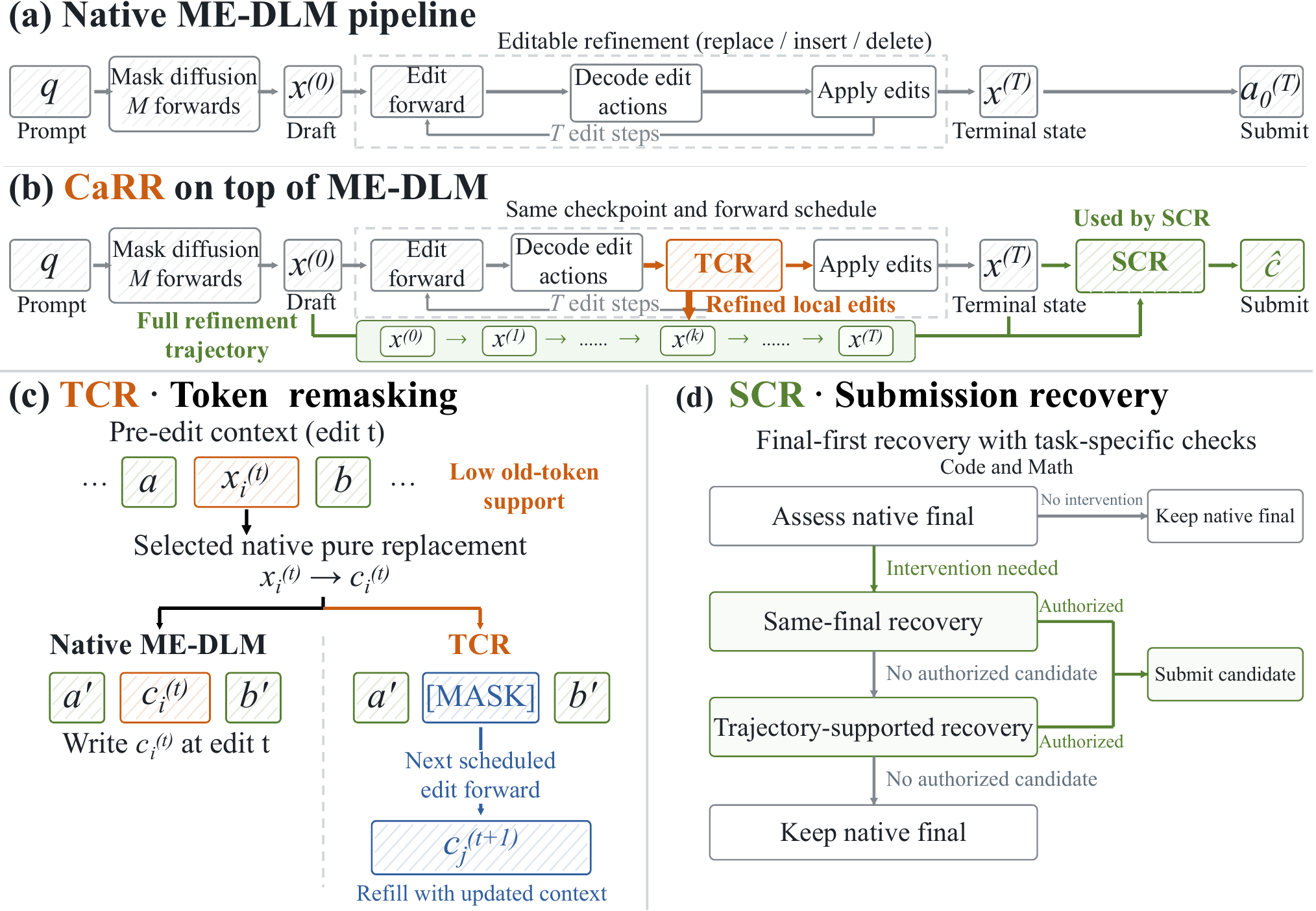}
    \vspace{-5mm}
    \caption{Overview of \method{} on top of ME-DLM. (a) Native decoding generates a draft $x^{(0)}$, applies $T$ scheduled edit updates, and submits the native final view $a_0^{(T)}$. (b) \method{} adds TCR during refinement and SCR at submission. (c) TCR remasks selected native replacements and refills them on the next scheduled forward using updated context. (d) SCR uses candidate views $C^{(t)}$ from the completed trajectory for final-first output recovery.}
    \label{fig:carr-overview}
    \vspace{-6mm}
\end{figure*}

Figure~\ref{fig:carr-overview} shows where \method{} intervenes. Token Commitment Remasking (\tokenguard) acts before a selected replacement is applied: it writes a mask instead of the proposed token and lets the next scheduled forward refill that position. The refill can therefore use surrounding changes that were unavailable to the original proposal. Submission Commitment Recovery (\trajsubmit) acts after the last edit: it extracts candidate views from the saved response states and uses task-specific output rules to determine the submission. The stages are connected through the trajectory: \tokenguard changes the states that \trajsubmit subsequently examines. Throughout, $x^{(t)}$ denotes the response state of the run under discussion.

Algorithm~\ref{alg:carr} summarizes the execution order of \method{}. TCR operates inside the native edit loop, whereas SCR is invoked once after the completed trajectory has been collected.

\begin{algorithm}[t]
\caption{\method{} inference}
\label{alg:carr}
\small
\begin{algorithmic}[1]
\Require Prompt $q$; frozen ME-DLM; budgets $(M,T)$; domain $d$
\Ensure Final submission $\hat c$

\State $x^{(0)} \gets \Call{MaskDiffusion}{q,M}$

\For{$t=0,\ldots,T-1$}
    \State $\mathcal{E}^{(t)}
    \gets \Call{EditForward}{q,x^{(t)}}$

    \State $\mathcal{E}^{(t)}
    \gets \Call{TCR}{x^{(t)},\mathcal{E}^{(t)}}$
    \Comment{refill and defer replacements}

    \State $x^{(t+1)}
    \gets \Call{ApplyEdits}{x^{(t)},\mathcal{E}^{(t)}}$
\EndFor

\State $C^{(0:T)}
\gets \Call{ExtractViews}{q,x^{(0:T)},d}$

\State $\hat c
\gets \Call{SCR}{C^{(0:T)},d}$
\Comment{final-first recovery}

\State \Return $\hat c$

\end{algorithmic}
\end{algorithm}
\vspace{-4mm}

\subsection{Token Commitment Remasking}
\label{sec:tcr}

An immediate replacement answers two questions at once: should the current token change, and what should replace it? Low support for the current token motivates changing it, but does not establish the reliability of the proposed substitute. Moreover, parallel replacements are predicted before neighboring edits are applied. \tokenguard separates these decisions by adapting the detect--then--remask principle of T2M~\citep{yao2026t2m} to native ME-DLM replacements.

\paragraph{Choosing which replacements to defer.}
At editable position $i$, let $x_i^{(t)}$ be the existing token and $c_i^{(t)}$ the native decoder's proposed token. \tokenguard considers only \emph{pure replacements}: native actions that replace an existing response token without inserting or deleting a token at that position. Among these actions, it identifies positions where the scheduled forward assigns low probability to the existing token:
\begin{equation}
    p_\theta^{\mathrm{cur}}\!\left(x_i^{(t)}\mid q,x^{(t)},i\right)
    <\tau_{\mathrm{lp}},
    \qquad \tau_{\mathrm{lp}}=0.30.
    \label{eq:tcr-score}
\end{equation}
Here $p_\theta^{\mathrm{cur}}$ is the current-token distribution of the frozen backbone. The probability in Equation~\ref{eq:tcr-score} measures support for the \emph{old token}, not confidence in $c_i^{(t)}$. Rather than replacing one uncertain commitment with another immediately, \tokenguard uses low old-token support to decide where to defer the replacement. The threshold is a frozen empirical operating point; Section~\ref{sec:selection-controls} evaluates selection signals separately from the deferred-refill operation using actual-generation controls.

\paragraph{Keeping remasking local and sparse.}
INSERT and DELETE actions can change token positions, and a remask must remain associated with the token being edited. \tokenguard therefore excludes positions within distance one of a native structural edit and allows each surviving token to be remasked at most once, carrying its remask history through position shifts. Positions involving control tokens are excluded, and new remasks are disabled on the last edit step, when no later forward remains to fill them. Among the remaining candidates, it selects the lowest-support positions under a per-step cap based on 25\% of all editable response positions, with a minimum cap of one when candidates exist. We denote the selected positions by $\mathcal R^{(t)}$. The numerical threshold, count limit, and ratio follow the frozen configuration adopted from T2M v2~\citep[Section~5.4]{yao2026t2m_v2}; Appendix~\ref{app:tcr-protocol} gives the exact eligibility conditions, rounding, and position-tracking rules.

\paragraph{Refilling after the context changes.}
For each $i\in\mathcal R^{(t)}$, \tokenguard writes $\mathtt{[MASK]}$ instead of $c_i^{(t)}$ and applies the other native actions for that step. The resulting state $x^{(t+1)}$ contains the pending mask together with the surrounding edits. Let $j$ be the mask's position in this updated state; INSERT/DELETE actions elsewhere may have shifted it from $i$. On the next scheduled forward, the refill is the highest-probability token under the current-token distribution, excluding $\mathtt{[PAD]}$, $\mathtt{[DEL]}$, and $\mathtt{[MASK]}$ from the maximization:
\begin{equation}
    c_j^{(t+1)}\leftarrow
    \arg\max_w p_\theta^{\mathrm{cur}}\!\left(w\mid q,x^{(t+1)},j\right).
    \label{eq:next-forward-refill}
\end{equation}
The refill is written to both aligned proposal slots so that native edit application treats it as a replacement rather than an insertion. It is then materialized in the subsequent edited response. The difference from immediate replacement is the conditioning context: $c_i^{(t)}$ was proposed from $x^{(t)}$, whereas the refill uses the surrounding edits already present in $x^{(t+1)}$. This deferral reuses the next scheduled forward instead of adding a separate model call. The exact proposal alignment and position mapping are specified in Appendix~\ref{app:tcr-protocol}.

\subsection{Submission Commitment Recovery}
\label{sec:scr}

Even after token refinement, a usable submission can be absent from the terminal response but present in an earlier state. \trajsubmit uses the completed trajectory to recover such candidates. Its \emph{final-first} policy keeps the native final unless the task's output contract identifies an actionable defect or consistency violation and supports a recovery. Thus, the trajectory provides evidence for a submission decision, rather than requiring another round of generation.

\paragraph{Extracting submission evidence.}
A response can contain several candidate views: for example, a Python block embedded in prose, or a mathematical answer stated in both a box and a concluding equation. \trajsubmit extracts these views from each saved response using fixed domain-specific rules. We write the ordered views from $x^{(t)}$ as $C^{(t)}$, and denote the native final view by $a_0^{(T)}$. Candidate views are not new model samples; they are alternative readings or permitted normalizations of text already generated during refinement.

\paragraph{Recovering in final-first order.}
\trajsubmit first assesses the native final using the domain contract. When that assessment calls for intervention, it tries recovery from the same final response before applying the domain's trajectory-supported recovery rules. The first authorized result becomes the submission $\hat c$; if neither stage resolves a candidate, the native final is retained. Both assessment and recovery may consult trajectory evidence. The concrete contracts below determine what counts as an actionable problem and which alternative can be returned; the formal decision rule appears in Appendix~\ref{app:scr-contracts}.

\paragraph{Code recovery.}
For Code, the contract checks whether a Python candidate parses under the AST and satisfies the public-interface requirement. A structurally valid native final is preserved. Otherwise, \trajsubmit first attempts a valid extraction from that same response, such as recovering a complete Python block from surrounding text. If this fails, it scans earlier states from nearest to farthest and returns the first candidate satisfying the same structural contract. This order distinguishes recovery within the terminal response from recovery of a program that survived only in an earlier state. These checks establish structural validity, not functional correctness.

\paragraph{Math recovery.}
For Math, surface differences can hide equivalent answers, while different parts of a response can support conflicting answers. \trajsubmit extracts boxed answers, explicit conclusions, terminal equations, and safe malformed-box views, then groups equivalent answers under frozen normalization rules. Agreement across distinct views within one state supplies support; repeated equivalent answers across consecutive states supply persistence. The recovery rules compare this support and persistence across competing answer groups to recover from an invalid final box or authorize narrowly specified terminal-consensus corrections. A separate canonicalization rule normalizes an explicit closed rational expression already present in the final box. A trajectory-supported candidate need not come from an earlier state: the terminal response can supply the candidate while earlier states supply evidence for it. Appendix~\ref{app:scr-contracts} specifies these branches, their thresholds, and their precedence.
The resolver uses the following selection precedence, subject to the frozen branch guards: for an invalid final box, same-final recovery $\to$ persistent earlier-box recovery $\to$ terminal-basin recovery; for a well-formed box, rational canonicalization $\to$ edit-local correction $\to$ answer-linked consensus. The first authorized branch returns its candidate; otherwise the native final is retained. Appendix~\ref{app:discussion-math} distinguishes rule roles and evaluates a four-branch subset. The evaluated adapters cover Code and Math. Extension to open-ended tasks requires a criterion for when recovery is warranted and an evaluation of whether it improves output quality.

The backbone parameters, response-canvas limit, and native $(M,T)$ allocation remain fixed. \tokenguard reuses scheduled forwards, while \trajsubmit performs deterministic CPU-side processing after generation. The scheduled NFE therefore remains $M+T$, although the edited response lengths can differ from native decoding.

\section{Experiments}
\label{sec:experiments}

\subsection{Experimental Setup}
\label{sec:exp-setup}
 
\paragraph{Datasets and Metrics.}
We evaluate \method{} on four benchmarks spanning code generation and mathematical reasoning. For code generation, we use HumanEval~\citep{chen2021codex} and MBPP~\citep{austin2021mbpp}, and report both the original Base accuracy and standard EvalPlus+ accuracy~\citep{liu2023evalplus}, where a completion must pass both the Base and additional EvalPlus tests. For mathematical reasoning, we use GSM8K~\citep{cobbe2021gsm8k} and MATH-500~\citep{hendrycks2021math} with benchmark-standard answer equivalence. Unless stated otherwise, ``average'' denotes the unweighted mean over these six metrics.

\paragraph{Baselines.}
We use ME-DLM~\citep{ren2026medlm} as the native decoder and compare against two locally evaluated methods under the same scheduled NFE: CoRe~\citep{zhai2026core}, which reserves part of the forward budget for perturbation probes, and a fixed-quota port of KLASS~\citep{kim2025klass}, which retains the native per-step commitment quota. Budget accounting for the same-NFE comparison is detailed in Appendix~\ref{app:budget-accounting}. For context, we also include the published per-benchmark scores of Soft Mask~\citep{hersche2025softmask}, EvoToken~\citep{zhong2026evotoken}, and LLaDA-Instruct~\citep{nie2025llada}, as reported in Table~2 of \citet{ren2026medlm}; these rows use their reported inference settings and are not included in paired analyses. The \tokenguard-only and \trajsubmit-only variants serve as component ablations, and Replace-Veto, which applies the same selection rule as \tokenguard but keeps the old token instead of remasking it, serves as the mechanism control.

\paragraph{Implementation Details.}
We evaluate the released ME-DLM Stage-3 model~\citep{ren2026medlm} at scheduled NFE budgets of 64, 128, 256, and 512 while preserving its native mask/edit allocation; exact schedules and baseline accounting are provided in Appendix~\ref{app:budget-accounting}. \method{} uses the frozen configurations of Sections~3.3--3.4. The Math SCR contract was developed using $1/8$ evidence and frozen thereafter; replacing its $1/8$ Math results with the corresponding TCR-only values still yields a 7.25 average gain over ME-DLM (Appendix~\ref{app:math-dev-sensitivity}). At $1/8$, matched warmed batch-one measurements including normal result writing show median paired latency increases of 0.75\% on HumanEval and 3.37\% on MATH-500 for Full \method{} over Native. Appendix~\ref{app:compute} reports the measurement boundary and observed GPU/CPU peak memory.

\subsection{Main Results}
\label{sec:main-results}

\begin{table}[t]
 \caption{Fixed-budget accuracy (\%) on code generation and mathematical
 reasoning. $\dagger$ denotes published results and $\ddagger$ local fixed-NFE
 ports. Average is the
 unweighted mean over six metrics. Gray bold text reports the absolute gain of \method{} over ME-DLM at the same budget. Cell shading indicates relative performance among the locally evaluated methods within each budget; darker shading indicates stronger performance.}
  \label{tab:main-results}
  \centering
  \setlength{\tabcolsep}{2.5pt}
  \renewcommand{\arraystretch}{1}
  \small
  \resizebox{\linewidth}{!}{%
  \begin{tabular}{ccccccccc}
  \toprule
  \multicolumn{1}{c}{Budget} & \multicolumn{1}{c}{Model} & \multicolumn{6}{c}{Evaluation Dataset} & \multicolumn{1}{c}{Average} \\
  \cmidrule(lr){3-8}
  & & HumanEval & HumanEval+ & MBPP & MBPP+ & GSM8K & MATH-500 & \\
  \midrule
    \multirow{6}{*}{$1/8$} & LLaDA-Instruct$^\dagger$ & 12.20 & 9.80 & 17.50 & 15.30 & 50.30 & 20.20 & 20.88 \\
  \cmidrule(lr){2-9}
     & ME-DLM & \cellcolor{carrred!6}23.17 & \cellcolor{carrred!6}20.73 & \cellcolor{carrred!6}26.46 & \cellcolor{carrred!6}23.28 & \cellcolor{carrred!10}63.99 & \cellcolor{carrred!22}33.80 & \cellcolor{carrred!6}31.90 \\
     & CoRe$^\ddagger$ & \cellcolor{carrred!19}25.61 & \cellcolor{carrred!24}23.78 & \cellcolor{carrred!8}26.98 & \cellcolor{carrred!6}23.28 & \cellcolor{carrred!6}63.31 & \cellcolor{carrred!6}32.40 & \cellcolor{carrred!10}32.56 \\
     & KLASS$^\ddagger$ & \cellcolor{carrred!35}28.66 & \cellcolor{carrred!48}28.05 & \cellcolor{carrred!23}30.95 & \cellcolor{carrred!17}25.93 & \cellcolor{carrred!7}63.53 & \cellcolor{carrred!13}33.00 & \cellcolor{carrred!24}35.02 \\
     & \method & \cellcolor{carrred!48}31.10 & \cellcolor{carrred!48}28.05 & \cellcolor{carrred!48}37.83 & \cellcolor{carrred!48}33.33 & \cellcolor{carrred!48}69.75 & \cellcolor{carrred!48}36.00 & \cellcolor{carrred!48}39.34 \\
    & \textcolor{black!65}{\textbf{\textit{vs. ME-DLM}}} & \textcolor{black!65}{\textbf{+7.93}} & \textcolor{black!65}{\textbf{+7.32}} & \textcolor{black!65}{\textbf{+11.37}} & \textcolor{black!65}{\textbf{+10.05}} & \textcolor{black!65}{\textbf{+5.76}} & \textcolor{black!65}{\textbf{+2.20}} & \textcolor{black!65}{\textbf{+7.44}} \\
  \midrule
    \multirow{8}{*}{$1/4$} & Soft Mask$^\dagger$ & 24.80 & 23.00 & 32.30 & -- & 62.30 & 19.80 & -- \\
     & EvoToken$^\dagger$ & -- & -- & -- & -- & 72.30 & 31.20 & -- \\
     & LLaDA-Instruct$^\dagger$ & 29.30 & 25.60 & 26.70 & 23.00 & 66.00 & 23.40 & 32.33 \\
  \cmidrule(lr){2-9}
     & ME-DLM & \cellcolor{carrorange!10}40.85 & \cellcolor{carrorange!11}38.41 & \cellcolor{carrorange!18}43.12 & \cellcolor{carrorange!17}37.30 & \cellcolor{carrorange!10}77.41 & \cellcolor{carrorange!11}42.40 & \cellcolor{carrorange!11}46.58 \\
     & CoRe$^\ddagger$ & \cellcolor{carrorange!6}40.24 & \cellcolor{carrorange!6}37.80 & \cellcolor{carrorange!21}43.92 & \cellcolor{carrorange!21}38.36 & \cellcolor{carrorange!6}77.18 & \cellcolor{carrorange!48}45.20 & \cellcolor{carrorange!15}47.12 \\
     & KLASS$^\ddagger$ & \cellcolor{carrorange!14}41.46 & \cellcolor{carrorange!20}39.63 & \cellcolor{carrorange!6}39.95 & \cellcolor{carrorange!6}34.66 & \cellcolor{carrorange!8}77.33 & \cellcolor{carrorange!6}42.00 & \cellcolor{carrorange!6}45.84 \\
     & \method & \cellcolor{carrorange!48}46.34 & \cellcolor{carrorange!48}43.29 & \cellcolor{carrorange!48}51.06 & \cellcolor{carrorange!48}44.71 & \cellcolor{carrorange!48}79.91 & \cellcolor{carrorange!32}44.00 & \cellcolor{carrorange!48}51.55 \\
    & \textcolor{black!65}{\textbf{\textit{vs. ME-DLM}}} & \textcolor{black!65}{\textbf{+5.49}} & \textcolor{black!65}{\textbf{+4.88}} & \textcolor{black!65}{\textbf{+7.94}} & \textcolor{black!65}{\textbf{+7.41}} & \textcolor{black!65}{\textbf{+2.50}} & \textcolor{black!65}{\textbf{+1.60}} & \textcolor{black!65}{\textbf{+4.97}} \\
  \midrule
    \multirow{8}{*}{$1/2$} & Soft Mask$^\dagger$ & 38.30 & 33.80 & 48.40 & -- & 79.40 & 38.80 & -- \\
     & EvoToken$^\dagger$ & -- & -- & -- & -- & 81.80 & 37.40 & -- \\
     & LLaDA-Instruct$^\dagger$ & 43.30 & 38.40 & 41.50 & 34.70 & 71.00 & 26.20 & 42.52 \\
  \cmidrule(lr){2-9}
     & ME-DLM & \cellcolor{carrblue!6}48.17 & \cellcolor{carrblue!6}44.51 & \cellcolor{carrblue!19}53.70 & \cellcolor{carrblue!6}45.24 & \cellcolor{carrblue!35}84.08 & \cellcolor{carrblue!32}48.00 & \cellcolor{carrblue!6}53.95 \\
     & CoRe$^\ddagger$ & \cellcolor{carrblue!6}48.17 & \cellcolor{carrblue!6}44.51 & \cellcolor{carrblue!27}54.50 & \cellcolor{carrblue!36}47.35 & \cellcolor{carrblue!6}82.18 & \cellcolor{carrblue!48}48.60 & \cellcolor{carrblue!11}54.22 \\
     & KLASS$^\ddagger$ & \cellcolor{carrblue!48}55.49 & \cellcolor{carrblue!48}51.83 & \cellcolor{carrblue!6}52.38 & \cellcolor{carrblue!14}45.77 & \cellcolor{carrblue!28}83.62 & \cellcolor{carrblue!6}47.00 & \cellcolor{carrblue!47}56.01 \\
     & \method & \cellcolor{carrblue!20}50.61 & \cellcolor{carrblue!24}47.56 & \cellcolor{carrblue!48}56.61 & \cellcolor{carrblue!48}48.15 & \cellcolor{carrblue!48}84.91 & \cellcolor{carrblue!48}48.60 & \cellcolor{carrblue!48}56.07 \\
    & \textcolor{black!65}{\textbf{\textit{vs. ME-DLM}}} & \textcolor{black!65}{\textbf{+2.44}} & \textcolor{black!65}{\textbf{+3.05}} & \textcolor{black!65}{\textbf{+2.91}} & \textcolor{black!65}{\textbf{+2.91}} & \textcolor{black!65}{\textbf{+0.83}} & \textcolor{black!65}{\textbf{+0.60}} & \textcolor{black!65}{\textbf{+2.12}} \\
  \midrule
    \multirow{8}{*}{$1/1$} & Soft Mask$^\dagger$ & 57.80 & 50.00 & 56.40 & -- & 84.00 & 41.40 & -- \\
     & EvoToken$^\dagger$ & -- & -- & -- & -- & 84.50 & 41.00 & -- \\
     & LLaDA-Instruct$^\dagger$ & 44.50 & 37.80 & 49.50 & 40.70 & 71.90 & 27.60 & 45.33 \\
  \cmidrule(lr){2-9}
     & ME-DLM & \cellcolor{carrpurple!31}56.10 & \cellcolor{carrpurple!20}52.44 & \cellcolor{carrpurple!11}58.47 & \cellcolor{carrpurple!6}49.47 & \cellcolor{carrpurple!6}84.91 & \cellcolor{carrpurple!6}50.00 & \cellcolor{carrpurple!6}58.56 \\
     & CoRe$^\ddagger$ & \cellcolor{carrpurple!6}54.27 & \cellcolor{carrpurple!6}51.83 & \cellcolor{carrpurple!6}58.20 & \cellcolor{carrpurple!31}50.26 & \cellcolor{carrpurple!45}85.75 & \cellcolor{carrpurple!48}52.20 & \cellcolor{carrpurple!14}58.75 \\
     & KLASS$^\ddagger$ & \cellcolor{carrpurple!48}57.32 & \cellcolor{carrpurple!48}53.66 & \cellcolor{carrpurple!48}60.32 & \cellcolor{carrpurple!48}50.79 & \cellcolor{carrpurple!10}84.99 & \cellcolor{carrpurple!14}50.40 & \cellcolor{carrpurple!48}59.58 \\
     & \method & \cellcolor{carrpurple!31}56.10 & \cellcolor{carrpurple!20}52.44 & \cellcolor{carrpurple!22}58.99 & \cellcolor{carrpurple!23}50.00 & \cellcolor{carrpurple!48}85.82 & \cellcolor{carrpurple!10}50.20 & \cellcolor{carrpurple!21}58.93 \\
    & \textcolor{black!65}{\textbf{\textit{vs. ME-DLM}}} & \textcolor{black!65}{\textbf{+0.00}} & \textcolor{black!65}{\textbf{+0.00}} & \textcolor{black!65}{\textbf{+0.52}} & \textcolor{black!65}{\textbf{+0.53}} & \textcolor{black!65}{\textbf{+0.91}} & \textcolor{black!65}{\textbf{+0.20}} & \textcolor{black!65}{\textbf{+0.36}} \\
  \bottomrule
  \end{tabular}}
  \vspace{-6mm}
\end{table}

\begin{figure}[t]
    \centering
    \includegraphics[width=\linewidth]{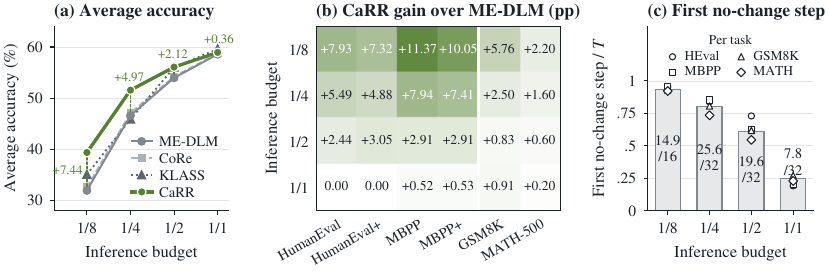}
    \vspace{-8mm}
    \caption{Performance across inference budgets.
    (a) Six-metric average accuracy of the locally evaluated methods.
    (b) Per-metric accuracy gain of \method{} over ME-DLM.
    (c) First no-change prediction step of native ME-DLM as a fraction of the
    scheduled $T$ (bars: pooled; markers: per task).}
    \label{fig:budget-overview}
    \vspace{-6mm}
\end{figure}

Table~\ref{tab:main-results} reports the four-budget comparison, and Figure~\ref{fig:budget-overview} summarizes it. We make three observations.

\textbf{(1) \method{} consistently improves the native decoder.} Across the four budgets, \method{} improves 22 of the 24 metric--budget settings over ME-DLM and matches the remaining two. Figure~\ref{fig:budget-overview}(b) shows gains across both code generation and mathematical reasoning, with the largest improvements on code. At $1/8$, the gains reach 11.37 and 10.05 on MBPP and MBPP+, respectively.

\textbf{(2) The benefit decreases as the inference budget increases.} The average gain shrinks monotonically with the budget, from 7.44 at $1/8$ to 0.36 at $1/1$ (Figure~\ref{fig:budget-overview}(a)). Figure~\ref{fig:budget-overview}(c) shows that the first no-change prediction occurs progressively earlier relative to the scheduled edit budget as the inference budget increases, reaching an average step of 7.8 out of 32 at
$1/1$. This pattern is consistent with reduced late-stage editing activity and fewer
opportunities for refinement and recovery. All scheduled edit forwards are still executed, and we use this statistic as a local-stationarity diagnostic. Per-task values are reported in Appendix~\ref{app:saturation}.

\textbf{(3) The advantage is broad at small budgets and less uniform at large budgets.} \method{} achieves or ties the best locally evaluated result on all six metrics at $1/8$ and achieves the best result on five of six at $1/4$. As the budget increases, the same-NFE baselines become more competitive: KLASS attains the highest average at $1/1$ and the best HumanEval results at $1/2$, while CoRe is best on MATH-500 at $1/4$ and $1/1$ and ties \method{} at $1/2$. Overall, \method{} provides its largest improvements under tight budgets and remains competitive at larger
ones.

\subsection{Ablation Study}
\label{sec:ablations}

\begin{table}[t]
\caption{Module ablation using the same six-metric average accuracy (\%) as
Table~\ref{tab:main-results}. Bold indicates the best result at each budget.}
\label{tab:module-ablation}
\centering
\small
\setlength{\tabcolsep}{7.5pt}
\begin{tabular}{lcccccc}
\toprule
Method & \tokenguard & \trajsubmit & $1/8$ & $1/4$ & $1/2$ & $1/1$ \\
\midrule
ME-DLM   &      &      & 31.90 & 46.58 & 53.95 & 58.56 \\
\tokenguard-only & \checkmark &      & 35.84 & 49.19 & 55.17 & 58.70 \\
\trajsubmit-only &      & \checkmark & 34.95 & 48.03 & 54.94 & 58.70 \\
\method  & \checkmark & \checkmark
         & \textbf{39.34} & \textbf{51.55}
         & \textbf{56.07} & \textbf{58.93} \\
\bottomrule
\end{tabular}
\vspace{-4mm}
\end{table}

To isolate the contribution of each module, we evaluate \tokenguard-only, which applies token remasking while retaining native final submission, and \trajsubmit-only which keeps native generation and applies submission recovery, under the same budgets. Table~\ref{tab:module-ablation} reports the six-metric averages. We make three observations.

\textbf{(1) Both modules are effective on their own.} \tokenguard-only and \trajsubmit-only improve the average accuracy over ME-DLM at all four budgets, showing that each stage provides useful gains on its own.

\textbf{(2) \tokenguard contributes more under tighter budgets.} \tokenguard-only exceeds \trajsubmit-only at $1/8$, $1/4$, and $1/2$, while the two become nearly identical at $1/1$.
This trend coincides with later first-no-change points under tighter budgets in Figure~\ref{fig:budget-overview}(c).

\textbf{(3) Combining the two modules provides additional gains.} Full \method{} achieves the highest average at every budget, exceeding the stronger single module by 3.50, 2.37, 0.90, and 0.22. At $1/8$ and $1/4$, the marginal gain of \trajsubmit after \tokenguard is 3.50 and 2.37, compared with 3.04 and 1.45 on the native trajectory. This interaction is not uniform at larger budgets. Task-level exceptions are reported in Appendix~\ref{app:four-arm}.

\textbf{Math rule simplification.}
Removing edit-local correction and answer-linked consensus leaves a four-branch SCR subset whose accuracy differs from full SCR by at most 0.61 on GSM8K and 0.20 on MATH-500 across both generation arms and all four budgets. Retained predicates and branch precedence are unchanged. Appendix~\ref{app:discussion-math} reports the full comparison on saved trajectories.

\subsection{Mechanism Analysis}
\label{sec:mechanism-analysis}

\begin{figure}[t]
    \centering
    \includegraphics[width=\linewidth]{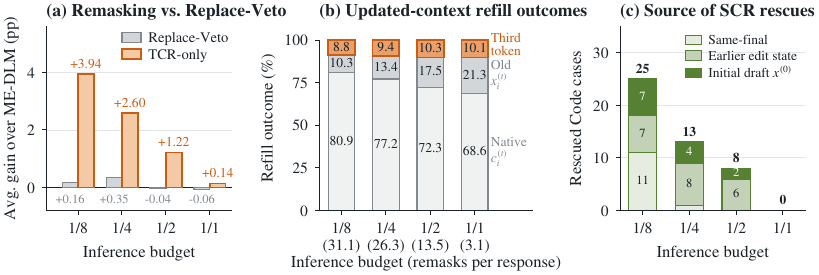}
    \vspace{-6mm}
    \caption{Mechanism analysis of \method{}.
    (a) Gain over ME-DLM of \tokenguard-only and the matched Replace-Veto control.
    (b) Outcome composition of updated-context refill at remasked positions.
    (c) Rescued Code Base cases of \trajsubmit by source state.}
    \label{fig:mechanism-analysis}
    \vspace{-6mm}
\end{figure}

We further analyze where the improvements come from. For \tokenguard, we compare against Replace-Veto and classify every updated-context refill in the \tokenguard-only runs by its outcome: the suppressed native replacement $c_i^{(t)}$, the old token $x_i^{(t)}$, or a third token. For \trajsubmit, we attribute each rescued Code Base case on pooled HumanEval and MBPP native ME-DLM trajectories to the state supplying the recovered candidate. Figure~\ref{fig:mechanism-analysis} shows the results, with remasks per response reported in parentheses in panel (b).

\textbf{(1) Remasking provides clear benefits beyond replacement veto.} Replace-Veto stays within 0.35 of ME-DLM at every budget, whereas \tokenguard-only gains 3.94 and 2.60 at $1/8$ and $1/4$, exceeding the matched control by 3.77 and 2.25 (Figure~\ref{fig:mechanism-analysis}(a)). These gaps support the contribution of updated-context refill beyond replacement veto under a shared selection and gating policy.

\textbf{(2) Updated-context refill changes the selected token outcome.} Figure~\ref{fig:mechanism-analysis}(b) shows that the refill differs from the suppressed native replacement in 19.1\%, 22.8\%, 27.7\%, and 31.4\% of remasked positions from $1/8$ to $1/1$. Across budgets, 10.3\%--21.3\% of remasks return to the old token, 8.8\%--10.3\% produce a third token, and remasking frequency decreases
from 31.1 to 3.1 positions per response.

\textbf{(3) Earlier trajectory states contribute substantially to \trajsubmit recovery.} Figure~\ref{fig:mechanism-analysis}(c) shows that earlier-state recovery provides 14, 12, and 8 of the 25, 13, and 8 rescued Code Base cases at $1/8$, $1/4$, and $1/2$, respectively; at $1/2$, same-final recovery adds no new case. Notably, 7, 4, and 2 rescues recover the initial mask-diffusion draft $x^{(0)}$. Because the Code resolver scans predecessor states from nearest
to farthest, these cases indicate that no later edit state supplied an authorized structurally valid recovery. No paired broken cases were observed in this analysis. Full Base and EvalPlus+ attribution results are reported in
Appendix~\ref{app:scr-analysis}.

\begingroup

\paragraph{Recovery simplification.}
\label{sec:worked-example}
At $1/8$, terminal submission and same-final extraction yield 146 and 154
Base+ passes on 542 TCR HumanEval/MBPP trajectories. History access raises
this to 172 with the same structural checks. The history-enabled policy reproduces the original Code SCR submissions in this replay. On \texttt{HumanEval/40}, SCR retrieves a working program from TCR state 15 after the final edit corrupts
it. For Math, boxed-only recovery improves accuracy by 0--0.20 percentage
points over terminal submission. On MATH-500, the four-branch subset
retains 1.00 of Full SCR's 1.20 percentage-point gain in each generation
arm. Appendix~\ref{app:discussion-math} reports the recovery variants and
their results. 

\subsection{Selection Controls}
\label{sec:selection-controls}

At $1/8$, Table~\ref{tab:gate-main} compares nine
generation policies on all 164 HumanEval and
500 MATH-500 questions under 64 scheduled forwards.
Terminal submissions evaluate the gate/refill policy
without SCR.

\begin{table}[H]

\centering
\small
\caption{Terminal-only accuracy (\%) for the gate controls.
HE+ requires the same completion to pass Base and
additional tests.}
\label{tab:gate-main}
\begin{tabular}{lrrr}
\toprule
Generation policy & HE & HE+ & MATH-500 \\
\midrule
Native                    & 23.17 & 20.73 & 33.80 \\
Original old-token        & 25.61 & 23.17 & 34.80 \\
No old threshold          & 28.05 & 26.22 & 35.20 \\
Proposed confidence       & 28.66 & 25.61 & 35.80 \\
Entropy                   & 29.27 & 26.83 & 35.20 \\
Random (seed 0)           & 25.61 & 23.78 & 35.00 \\
Old + proposed rank       & 26.22 & 23.78 & 35.80 \\
Old + entropy rank        & 26.22 & 23.78 & 35.60 \\
Old \& proposed $p<0.50$   & 25.61 & 23.17 & 35.40 \\
\bottomrule
\end{tabular}
\end{table}

All eight TCR variants improve observed terminal accuracy over Native on all three metrics. These results support deferred refill across the tested selection signals; the submitted old-token gate remains a fixed empirical operating point. Appendix~\ref{app:discussion-gates} reports the policy definitions, matched-count controls, intervention counts and complete terminal/original-SCR results.

\endgroup

\section{Conclusion}
\label{sec:conclusion}

We presented \method{}, a training-free framework for improving low-budget editable diffusion inference by coordinating token-level and submission-level commitments. TCR defers selected native pure replacements and refills them on the next scheduled edit forward using updated context, while SCR uses the resulting refinement trajectory for final-first output recovery. Across code generation and mathematical reasoning, \method{} improves average accuracy over native ME-DLM at all four tested budgets, with the largest gains under tight budgets and no additional backbone forwards. Ablations and mechanism analyses further show complementary benefits from updated-context refill and trajectory-based recovery. These results suggest that low-budget editable refinement can benefit from treating both local replacement and final submission as explicit commitment decisions.

\bibliographystyle{carr_refs}
\bibliography{references}

\appendix

\paragraph{Roadmap.}
Appendix~\ref{app:method-details} gives the exact inference protocols for \tokenguard and \trajsubmit.
Appendix~\ref{app:exp-details} documents budget accounting, evaluation metrics, and development protocols.
Appendix~\ref{app:four-arm} reports the full task-level component results.
Appendix~\ref{app:mechanism-details} expands the mechanism analyses in Figure~\ref{fig:mechanism-analysis}.
Appendix~\ref{app:saturation} provides the native first-no-change statistics used in Figure~\ref{fig:budget-overview}(c), and Appendix~\ref{app:compute} summarizes compute accounting.

\section{Detailed Inference Protocols}
\label{app:method-details}
All inference-time decisions use public task inputs, including Code entry-point metadata, model outputs, and frozen deterministic rules.
Benchmark labels, unit-test outcomes, and reference answers are used only for evaluation after a submission has been materialized.

\subsection{Exact TCR Protocol}
\label{app:tcr-protocol}
\tokenguard operates on the native edit actions decoded for a scheduled edit forward. At edit step $t$, let $x_i^{(t)}$ be the existing token at editable response position $i$, and let $c_i^{(t)}$ and $n_i^{(t)}$ denote the native current- and next-token proposals aligned to that destination position. A \emph{pure replacement} satisfies $c_i^{(t)}\neq x_i^{(t)}$ and $c_i^{(t)}=n_i^{(t)}$, with no length-changing action at that position. The next-token proposal is destination-aligned as in the runtime: the raw next-token head from the preceding position is shifted to the token position it predicts.

Following T2M's LowProb detector, the old-token gate is
\begin{equation}
    d_i^{(t)}=
    \mathbf 1\!\left\{
    p_\theta^{\mathrm{cur}}\!\left(x_i^{(t)}\mid q,x^{(t)},i\right)
    <\tau_{\mathrm{lp}}
    \right\}.
    \label{eq:app-tcr-detector}
\end{equation}
The probability in Equation~\ref{eq:app-tcr-detector} measures support for retaining the existing token; it is not confidence in the proposed replacement.

Let $m_i^{(t)}$ be the previous-remask count attached to the token lineage currently occupying position $i$, propagated through native INSERT/DELETE updates, and let $\delta_i^{(t)}$ be its distance to the nearest native structural edit. With
$\mathcal S=\{\mathtt{[PAD]},\mathtt{[DEL]},\mathtt{[MASK]},\mathtt{[EOS]}\}$,
the exact eligible and selected sets are
\begin{align}
\mathcal R_0^{(t)}
&=\Bigl\{i:\; i\text{ editable},\;d_i^{(t)}=1,\;
 c_i^{(t)}\neq x_i^{(t)},\;c_i^{(t)}=n_i^{(t)}, \nonumber\\
&\hspace{20mm}x_i^{(t)},c_i^{(t)}\notin\mathcal S,\;
 m_i^{(t)}<C_{\max},\;\delta_i^{(t)}>r,\;t<T-1\Bigr\},\label{eq:app-tcr-eligible}\\
k_t
&=\begin{cases}
0,&\mathcal R_0^{(t)}=\varnothing,\\
\min\!\left\{|\mathcal R_0^{(t)}|,
\max\!\left(1,\left\lfloor\rho_{\max}n_t\right\rfloor\right)\right\},&\text{otherwise},
\end{cases}\label{eq:app-tcr-cap}\\
\mathcal R^{(t)}
&=\operatorname{BottomK}_{i\in\mathcal R_0^{(t)}}
\!\left(p_\theta^{\mathrm{cur}}\!\left(x_i^{(t)}\mid q,x^{(t)},i\right);k_t\right).
\label{eq:app-tcr-select}
\end{align}
Here $n_t$ is the number of editable response positions. Equal support values retain native position order. Thus, TCR intervenes only on native pure replacements, remains isolated from length-changing edits, remasks each surviving lineage at most once, and disables new remasks on the terminal edit step.

\begin{table}[H]
\caption{Frozen \tokenguard configuration used in the original main-result runs; Appendix~\ref{app:discussion-gates} defines the additional selection controls.}
\label{tab:tcr-config}
\centering
\small
\begin{tabular}{lccl}
\toprule
Component & Symbol & Value & Role \\
\midrule
Retention threshold & $\tau_{\mathrm{lp}}$ & 0.30 & Select low-support old tokens \\
Per-step ratio cap & $\rho_{\max}$ & 0.25 & Cap remasked positions per edit step \\
Structural radius & $r$ & 1 & Isolate remasking from INSERT/DELETE actions \\
Lineage budget & $C_{\max}$ & 1 & Remask each surviving lineage at most once \\
Terminal remasking & -- & disabled & Require a later scheduled refill forward \\
Additional backbone forwards & -- & 0 & Reuse the next scheduled edit forward \\
\bottomrule
\end{tabular}
\end{table}

\paragraph{Updated-context refill.}
For each $i\in\mathcal R^{(t)}$, TCR writes $\mathtt{[MASK]}$ instead of the native concrete replacement and applies the other native actions for that step. After structural updates, let $j=\pi_t(i)$ be the position of the surviving masked lineage in $x^{(t+1)}$. On the next scheduled edit forward, the current-token logits provide the refill:
\begin{align}
 c_j^{(t+1)}
 &\leftarrow
 \arg\max_{w\notin\{\mathtt{[PAD]},\mathtt{[DEL]},\mathtt{[MASK]}\}}
 p_\theta^{\mathrm{cur}}\!\left(w\mid q,x^{(t+1)},j\right),\nonumber\\
 n_j^{(t+1)}&\leftarrow c_j^{(t+1)}.
 \label{eq:app-tcr-refill}
\end{align}
Writing the same refill to the aligned current- and next-token proposal slots preserves the native interpretation of the action as a replacement. The refill is then materialized by the native edit application. No additional backbone call is introduced.

\paragraph{Parameter selection.}
We use $(\tau_{\mathrm{lp}},C_{\max},\rho_{\max})=(0.30,1,0.25)$, following the LowProb operating point identified as most cost-efficient in T2M v2~\citep[Section~5.4 and Appendix~D]{yao2026t2m_v2}, where cost counts token modifications per sample. In that version, the LowProb detector sweeps $\tau_{\mathrm{lp}}\in\{0.1,0.3,0.5,0.7,0.9\}$, together with $C_{\max}\in\{1,3,5\}$ and $\rho_{\max}\in\{0.25,0.50,1.0\}$. Our fixed-budget ME-DLM adaptation keeps this numerical operating point and adds lineage tracking, structural isolation with $r=1$, and the finite-schedule terminal constraint above. The configuration was fixed before the reported benchmark runs, and the scheduled backbone budget remains $M+T$.

\paragraph{Development protocol.}
The frozen \tokenguard gate was validated on an auxiliary 256-request development set containing 64 Code and 192 Math requests, partitioned into 176 calibration requests and 80 held-out-source requests. Trajectory generation used the request inputs alone; clean responses and target token IDs were attached offline for analysis after generation. The detector and remasking policy were then fixed for the benchmark runs reported in the paper.

\subsection{Exact SCR Contracts}
\label{app:scr-contracts}
For task domain $d$, the deterministic adapter produces an ordered candidate-view sequence
\begin{equation}
 C^{(t)}=A_d\!\left(q,x^{(t)}\right)
 =\left(a_0^{(t)},\ldots,a_{J_t}^{(t)}\right),
 \qquad
 V_d\!\left(q,a;C^{(0:T)}\right)
 \in\{\mathrm{PASS},\mathrm{FAIL},\mathrm{ABSTAIN}\}.
 \label{eq:app-scr-status}
\end{equation}
Here $a_0^{(T)}$ is the native final view. The validator states are mechanical intervention statuses: PASS means no intervention is required, FAIL identifies an actionable mechanical defect or frozen consistency violation, and ABSTAIN indicates insufficient evidence for intervention.

A single resolver $R_d$ preserves the implemented recovery precedence: when the native final is marked FAIL, it first attempts same-final recovery and then applies the domain's trajectory-supported recovery rules. Let $a^\star=R_d(q;C^{(0:T)})$, with $a^\star=\bot$ when no recovery is authorized. The final-first submission rule is
\begin{equation}
\hat c=
\begin{cases}
 a^\star,&V_d\!\left(q,a_0^{(T)};C^{(0:T)}\right)=\mathrm{FAIL}
 \ \land\ a^\star\neq\bot,\\
 a_0^{(T)},&\text{otherwise}.
\end{cases}
\label{eq:app-scr-policy}
\end{equation}
Both recovery stages may use evidence from the complete trajectory. \emph{Trajectory-supported} describes the evidence used by a resolver, whereas \emph{earlier-state} describes the source of a returned candidate; these are not interchangeable.

\paragraph{Code candidate extraction.}
For the reported HumanEval and MBPP runs, the public entry point is taken from benchmark-supplied interface metadata after checking consistency with the public prompt. Prompt-based entry-point inference is used only in auxiliary workflows. The structural extractor considers complete Python fenced blocks, adjacent fence-marker segments that recover a complete block from malformed fencing, contiguous code-looking regions, and the raw response when no fence occurs. A candidate is structurally valid when it parses under the Python AST and contains the required public top-level definition.

When a state contains multiple structurally valid candidates, the resolver selects the latest valid structural candidate. The output precedence is fixed as follows:
\begin{enumerate}
    \item preserve the native final byte-for-byte when it is structurally valid;
    \item otherwise use a valid structural extraction from the same final state;
    \item otherwise scan predecessor states from nearest to farthest and use the first structurally valid candidate;
    \item if no candidate is authorized, return the native final unchanged.
\end{enumerate}
Program execution is reserved for offline evaluation; the selection rule uses parsing and public-interface checks only.

\paragraph{Math candidate views.}
The Math adapter extracts four surface families: a well-formed boxed answer, an explicit conclusion, a terminal equation, and a conservatively extracted malformed trailing box. Box, conclusion, and terminal-equation views are grouped into equivalence classes using frozen normalization and equivalence rules. Within a state, a unique maximum-support class defines the observed answer; tied classes abstain. Across states, consecutive equivalent observations form an answer basin characterized by run length and independent-view support.

\paragraph{Math recovery from an invalid final box.}
When the final box is invalid, the same-final resolver searches malformed-box, conclusion, and terminal-equation views and returns a candidate when one equivalence class has unique maximum support from at least two views. The trajectory resolver next considers earlier runs of a well-formed boxed answer with run length at least two. It prioritizes longer runs and breaks equal-length ties by the latest ending state, followed by the latest starting state. A terminal multi-view basin is eligible when it reaches the terminal state, persists for at least two observed states, has a safe candidate surface, and strictly exceeds every incompatible basin in both run length and total independent-view support.

\paragraph{Canonicalization and terminal consensus.}
For a well-formed final box, the resolver first applies exact canonicalization when the box already contains an eligible explicit closed rational expression. The expression must contain no free symbols, simplify to a rational number, change the surface representation, and satisfy the safe-box predicate.

If canonicalization does not apply, the resolver can use an edit-local dominant terminal-consensus correction when the terminal answer basin is persistent and strictly dominant, the terminal state supplies at least two independent supporting views, the normalized final and candidate scalar surfaces have character-level Levenshtein distance one or two, the candidate is nonzero, and the candidate surface is safe. If neither branch applies, the inherited answer-linked consensus resolver remains available as a fallback; that separate branch requires the conflicting final answer to participate in an invalid terminal arithmetic relation and to lack valid executable support. These are alternative recovery branches in both the 17-state implementation and its higher-budget generalization; the arithmetic checks are not prerequisites for the edit-local branch. Evidence not resolved by the frozen precedence and authorization rules retains the native final.

\paragraph{Frozen cross-budget transfer.}
The $1/8$ trajectory contains 17 complete response states and was used during development of the Math contract. The $1/4$, $1/2$, and $1/1$ trajectories contain 33 states. For the higher-budget runs, the terminal-state index is generalized while the absolute predicates remain fixed: persistence requires at least two consecutive states, dominance is strict in both run length and independent support, and the edit-local condition remains a one-to-two-character difference. The same thresholds are used across all higher-budget trajectories.

\section{Evaluation Protocol and Accounting}
\label{app:exp-details}

\subsection{Budget and Baseline-Port Accounting}
\label{app:budget-accounting}
All local comparisons are indexed by scheduled neural function evaluations (NFE).
ME-DLM, \tokenguard, \trajsubmit, Replace-Veto, fixed-quota KLASS, and full \method{} use the native mask/edit allocation.
CoRe reserves part of the same total NFE budget for perturbation probes.
Table~\ref{tab:budget-allocation} gives the exact schedules.

\begin{table}[H]
\caption{Scheduled forward allocation for the locally evaluated methods. ``Normal mask'' excludes CoRe perturbation probes.}
\label{tab:budget-allocation}
\centering
\small
\resizebox{\linewidth}{!}{%
\begin{tabular}{cccccc}
\toprule
Budget & Native $M$ & Native $T$ & Native/\method{} NFE & CoRe normal mask + probes + edit & Total NFE \\
\midrule
$1/8$ & 48 & 16 & 64  & $45+3+16$  & 64 \\
$1/4$ & 96 & 32 & 128 & $90+6+32$  & 128 \\
$1/2$ & 224 & 32 & 256 & $211+13+32$ & 256 \\
$1/1$ & 480 & 32 & 512 & $452+28+32$ & 512 \\
\bottomrule
\end{tabular}}
\end{table}

All local runs use the released ME-DLM Stage-3 model, response canvas $L=512$, and a single RTX~5090.
Our KLASS port keeps the native per-step commitment quota and applies the KLASS selection rule within that quota.
Soft Mask, EvoToken, and LLaDA-Instruct are included only as published context in Table~\ref{tab:main-results}; local paired analyses use the reproduced ME-DLM trajectories and locally evaluated methods.

\subsection{Benchmarks, Metrics, and Provenance}
\label{app:metric-provenance}

\begin{table}[H]
\caption{Evaluation sets and metrics used in the paper.}
\label{tab:benchmark-summary}
\centering
\small
\begin{tabular}{lcrl}
\toprule
Domain & Benchmark & Examples & Reported metrics \\
\midrule
Code & HumanEval & 164 & Base, EvalPlus+ \\
Code & MBPP (EvalPlus) & 378 & Base, EvalPlus+ \\
Math & GSM8K & 1,319 & Answer accuracy \\
Math & MATH-500 & 500 & Answer accuracy \\
\bottomrule
\end{tabular}
\end{table}

For Code, we report Base accuracy and standard EvalPlus+ accuracy, where the same completion must pass both the Base and additional EvalPlus tests. The paired SCR analyses use this same EvalPlus+ criterion.
The six-metric average in the main paper is the unweighted mean of HumanEval Base, HumanEval+, MBPP Base, MBPP+, GSM8K, and MATH-500 accuracy.
Locally evaluated main-table values are derived from the unified cross-method evaluation outputs; the four-arm comparison and rescue attribution use per-example paired evaluation records. Execution traces are used for refill-outcome and first-no-change analyses.

\subsection{Sensitivity to Math Development Evidence}
\label{app:math-dev-sensitivity}
The Math contract of \trajsubmit was developed using $1/8$ GSM8K and MATH-500 evidence and then frozen before transfer to $1/4$, $1/2$, and $1/1$.
To quantify the contribution of this development contact to the headline $1/8$ result, we replace the $1/8$ GSM8K and MATH-500 entries of full \method{} with their \tokenguard-only values from Table~\ref{tab:four-arm-full} (69.83 and 34.80), leaving the four Code metrics unchanged.
The resulting six-metric average is 39.16, corresponding to an absolute gain of 7.25 over native ME-DLM, compared with the reported gain of 7.44.
The higher-budget results use the same benchmark problem sets and the frozen contract without budget-specific retuning; they assess cross-budget transfer rather than generalization to an independent held-out dataset.

\section{Full Component Results}
\label{app:four-arm}
Table~\ref{tab:four-arm-full} expands the average ablation in Table~\ref{tab:module-ablation} into task-level results.
For Code, the ``+'' rows use the same standard EvalPlus+ criterion as the cross-method main table.

\begin{table}[H]
\caption{Full task-level component results (\%). ``SCR'' denotes native ME-DLM generation with \trajsubmit; ``TCR'' denotes \tokenguard generation with native final-only submission. Code ``+'' denotes standard EvalPlus+ accuracy.}
\label{tab:four-arm-full}
\centering
\small
\begin{tabular}{cclrrrr}
\toprule
Budget & Domain & Task / metric & ME-DLM & SCR & TCR & \method{} \\
\midrule
\multirow{6}{*}{$1/8$}
& Math & GSM8K & 63.99 & 64.90 & \textbf{69.83} & 69.75 \\
& Math & MATH-500 & 33.80 & 35.00 & 34.80 & \textbf{36.00} \\
& Code & HumanEval Base & 23.17 & 26.22 & 25.61 & \textbf{31.10} \\
& Code & HumanEval+ & 20.73 & 23.78 & 23.17 & \textbf{28.05} \\
& Code & MBPP Base & 26.46 & 31.75 & 33.07 & \textbf{37.83} \\
& Code & MBPP+ & 23.28 & 28.04 & 28.57 & \textbf{33.33} \\
\midrule
\multirow{6}{*}{$1/4$}
& Math & GSM8K & 77.41 & 77.56 & 79.53 & \textbf{79.91} \\
& Math & MATH-500 & 42.40 & 43.20 & 43.00 & \textbf{44.00} \\
& Code & HumanEval Base & 40.85 & 42.07 & 43.90 & \textbf{46.34} \\
& Code & HumanEval+ & 38.41 & 39.63 & 40.85 & \textbf{43.29} \\
& Code & MBPP Base & 43.12 & 46.03 & 46.56 & \textbf{51.06} \\
& Code & MBPP+ & 37.30 & 39.68 & 41.27 & \textbf{44.71} \\
\midrule
\multirow{6}{*}{$1/2$}
& Math & GSM8K & 84.08 & 84.31 & 84.61 & \textbf{84.91} \\
& Math & MATH-500 & 48.00 & \textbf{48.60} & 48.00 & \textbf{48.60} \\
& Code & HumanEval Base & 48.17 & 49.39 & \textbf{50.61} & \textbf{50.61} \\
& Code & HumanEval+ & 44.51 & 45.73 & \textbf{47.56} & \textbf{47.56} \\
& Code & MBPP Base & 53.70 & 55.29 & 54.23 & \textbf{56.61} \\
& Code & MBPP+ & 45.24 & 46.30 & 46.03 & \textbf{48.15} \\
\midrule
\multirow{6}{*}{$1/1$}
& Math & GSM8K & 84.91 & 85.14 & 85.60 & \textbf{85.82} \\
& Math & MATH-500 & 50.00 & \textbf{50.60} & 49.60 & 50.20 \\
& Code & HumanEval Base & \textbf{56.10} & \textbf{56.10} & \textbf{56.10} & \textbf{56.10} \\
& Code & HumanEval+ & \textbf{52.44} & \textbf{52.44} & \textbf{52.44} & \textbf{52.44} \\
& Code & MBPP Base & 58.47 & 58.47 & 58.73 & \textbf{58.99} \\
& Code & MBPP+ & 49.47 & 49.47 & 49.74 & \textbf{50.00} \\
\bottomrule
\end{tabular}
\end{table}

Full \method{} achieves the strongest six-metric average at every budget in Table~\ref{tab:module-ablation}, while Table~\ref{tab:four-arm-full} exposes localized task-level interactions between the two stages.
At $1/8$, \tokenguard-only is one GSM8K example higher than full \method{} (69.83 vs.\ 69.75); at $1/1$, \trajsubmit-only is two MATH-500 examples higher (50.60 vs.\ 50.20); and \tokenguard-only reaches 49.60 on $1/1$ MATH-500 compared with 50.00 for native ME-DLM.
These task-level results complement the aggregate module analysis in Section~\ref{sec:ablations}.

\section{Additional Mechanism Analysis}
\label{app:mechanism-details}

\subsection{TCR Refill Outcomes}
\label{app:refill-outcomes}
For every position remasked by \tokenguard, we pair the refill written on the next scheduled edit forward with the old token $x_i^{(t)}$ and the suppressed native replacement $c_i^{(t)}$ recorded at the remasking step.
The runtime stores selected positions and refill tokens in the same order, and refill counts match selection counts for every paired step in the benchmark logs.
Table~\ref{tab:refill-outcomes} reports the outcome composition by task, while Figure~\ref{fig:mechanism-analysis}(b) shows the pooled composition.

\begin{table}[H]
\caption{Updated-context refill outcomes (\%) by task. ``Native'' denotes a refill equal to the suppressed native replacement $c_i^{(t)}$, ``Old'' a refill equal to $x_i^{(t)}$, and ``Third'' any other token. Remasks per response are pooled across all 2,361 responses at each budget.}
\label{tab:refill-outcomes}
\centering
\scriptsize
\resizebox{\linewidth}{!}{%
\begin{tabular}{cc|ccc|ccc|ccc|ccc}
\toprule
& Remasks & \multicolumn{3}{c|}{HumanEval} & \multicolumn{3}{c|}{MBPP} & \multicolumn{3}{c|}{GSM8K} & \multicolumn{3}{c}{MATH-500} \\
Budget & / resp. & Native & Old & Third & Native & Old & Third & Native & Old & Third & Native & Old & Third \\
\midrule
$1/8$ & 31.1 & 79.4 & 9.6 & 10.9 & 80.2 & 9.6 & 10.3 & 81.3 & 10.5 & 8.2 & 81.1 & 10.6 & 8.3 \\
$1/4$ & 26.3 & 77.3 & 12.2 & 10.6 & 77.2 & 11.8 & 11.0 & 77.1 & 14.0 & 8.9 & 77.5 & 13.6 & 8.9 \\
$1/2$ & 13.5 & 74.4 & 14.6 & 11.0 & 74.4 & 15.7 & 9.9 & 72.0 & 18.0 & 10.0 & 71.0 & 18.1 & 10.9 \\
$1/1$ & 3.1 & 71.1 & 16.1 & 12.8 & 69.9 & 20.5 & 9.5 & 69.6 & 20.9 & 9.5 & 66.0 & 22.9 & 11.1 \\
\bottomrule
\end{tabular}}
\end{table}

Across tasks, the updated-context refill frequently reproduces the suppressed native replacement and also produces two systematic alternatives: returning to the old token and selecting a third token.
Pooled across tasks, the refill differs from the suppressed native replacement in 19.1\%, 22.8\%, 27.7\%, and 31.4\% of remasked positions from $1/8$ to $1/1$.
The remasking frequency decreases over the same budgets, from 31.1 to 3.1 positions per response.

\subsection{SCR Paired Attribution}
\label{app:scr-analysis}
The three-arm Code analysis reuses identical saved native ME-DLM trajectories and compares the native final, same-final structural recovery, and full \trajsubmit with earlier-state recovery.
Table~\ref{tab:scr-paired-full} reports Base and EvalPlus+ outcomes pooled over HumanEval and MBPP ($164+378=542$ examples).

\begin{table}[H]
\caption{Paired Code attribution for \trajsubmit. R/B denotes rescued/broken cases relative to the preceding arm.}
\label{tab:scr-paired-full}
\centering
\small
\setlength{\tabcolsep}{3pt}
\begin{tabular}{ccrrrrr}
\toprule
Metric & Budget & Native final & Same-final & Full SCR & Native$\rightarrow$same R/B & Same$\rightarrow$full R/B \\
\midrule
\multirow{4}{*}{Base}
& $1/8$ & 138 & 149 & 163 & 11/0 & 14/0 \\
& $1/4$ & 230 & 231 & 243 & 1/0 & 12/0 \\
& $1/2$ & 282 & 282 & 290 & 0/0 & 8/0 \\
& $1/1$ & 313 & 313 & 313 & 0/0 & 0/0 \\
\midrule
\multirow{4}{*}{EvalPlus+}
& $1/8$ & 122 & 132 & 145 & 10/0 & 13/0 \\
& $1/4$ & 204 & 205 & 215 & 1/0 & 10/0 \\
& $1/2$ & 244 & 244 & 250 & 0/0 & 6/0 \\
& $1/1$ & 273 & 273 & 273 & 0/0 & 0/0 \\
\bottomrule
\end{tabular}
\end{table}

The $1/2$ setting provides the clearest separation between same-final and earlier-state recovery: same-final extraction leaves both Base and EvalPlus+ counts unchanged, while earlier-state recovery adds eight Base and six EvalPlus+ passes.
Across the paired Code analysis, no broken case is observed in the transitions reported in Table~\ref{tab:scr-paired-full}.

\subsection{SCR Rescue Sources}
\label{app:scr-sources}
Table~\ref{tab:scr-sources} decomposes rescued cases by the state that supplies the returned candidate.
These analyses reuse native ME-DLM trajectories; under the unified notation, the initial draft is $x^{(0)}$ and the intermediate edit states are $x^{(1)},\ldots,x^{(T-1)}$.

\begin{table}[H]
\caption{Source of rescued Code cases on pooled HumanEval and MBPP native ME-DLM trajectories.}
\label{tab:scr-sources}
\centering
\small
\setlength{\tabcolsep}{2pt}
\begin{tabular}{ccrrrr}
\toprule
Metric & Budget & Same-final & Earlier edit & Initial draft $x^{(0)}$ & Total \\
\midrule
\multirow{4}{*}{Base}
& $1/8$ & 11 & 7 & 7 & 25 \\
& $1/4$ & 1 & 8 & 4 & 13 \\
& $1/2$ & 0 & 6 & 2 & 8 \\
& $1/1$ & 0 & 0 & 0 & 0 \\
\midrule
\multirow{4}{*}{EvalPlus+}
& $1/8$ & 10 & 7 & 6 & 23 \\
& $1/4$ & 1 & 7 & 3 & 11 \\
& $1/2$ & 0 & 4 & 2 & 6 \\
& $1/1$ & 0 & 0 & 0 & 0 \\
\bottomrule
\end{tabular}
\end{table}

Earlier states provide most of the additional Code recovery at the tighter budgets.
At $1/2$, all eight rescued Base cases come from earlier states, including two from the initial mask-diffusion draft.
Selection of $x^{(0)}$ occurs after all later edit states have been evaluated without an authorized structurally valid candidate.

\section{Native First-No-Change Analysis}
\label{app:saturation}
To characterize when native refinement first becomes locally stationary, we use the logged \texttt{real\_edit\_step}, defined as the first edit step at which the predicted response no longer changes, capped at the scheduled edit budget $T$.
All scheduled edit forwards are executed; this statistic records the first unchanged prediction within that fixed schedule.
Table~\ref{tab:real-edit-steps} reports task-level averages and the example-weighted pooled value.

\begin{table}[H]
\caption{Average first no-change prediction step of native ME-DLM. ``Pooled'' is example-weighted across the four benchmarks; the denominator is the scheduled edit budget $T$.}
\label{tab:real-edit-steps}
\centering
\small
\begin{tabular}{cccccc}
\toprule
Budget & HumanEval & MBPP & GSM8K & MATH-500 & Pooled \\
\midrule
$1/8$ & 15.29 & 15.27 & 14.87 & 14.75 & 14.94 / 16 \\
$1/4$ & 26.85 & 27.28 & 25.81 & 23.50 & 25.63 / 32 \\
$1/2$ & 23.35 & 20.07 & 19.76 & 17.43 & 19.57 / 32 \\
$1/1$ & 6.07 & 6.48 & 8.64 & 7.27 & 7.83 / 32 \\
\bottomrule
\end{tabular}
\end{table}

Relative to the scheduled edit budget, the first unchanged prediction occurs progressively earlier as the inference budget increases: the pooled fraction is 0.933, 0.801, 0.611, and 0.245 from $1/8$ to $1/1$.

This trend accompanies lower remasking activity and smaller average gains of \method{} at larger budgets, consistent with reduced late-stage editing activity. The corresponding pooled first-no-change trend is summarized in Figure~\ref{fig:budget-overview}(c).

\section{Compute Accounting}
\label{app:compute}
Local comparisons use scheduled NFE as the common inference-budget measure.
For ME-DLM, \tokenguard, \trajsubmit, Replace-Veto, fixed-quota KLASS, and full \method{}, the scheduled backbone budget is the native $M+T$ allocation.
\tokenguard reads logits and decoded actions from scheduled edit forwards and performs refill on the next scheduled forward; \trajsubmit performs deterministic CPU-side candidate extraction and resolution after generation.
Full \method{} therefore adds no backbone forward to the native schedule.

Actual runtime cost also depends on the response lengths produced along the edit trajectory, which determine processed token positions and attention cost.
CoRe allocates a portion of its matched total NFE to perturbation probes as detailed in Table~\ref{tab:budget-allocation}.
Scheduled NFE is therefore the controlled inference-budget axis throughout the paper. The following matched measurements report request latency and GPU/CPU memory, including necessary history, decoding, selection and normal result serialization/write.
\begingroup
\begin{table}[H]

\centering
\small
\setlength{\tabcolsep}{3pt}
\begin{tabular}{llrrrrr}
\toprule
Task & Arm & Median (s) & p95 (s) & Alloc. & Reserved & CPU RSS \\
& & & & \multicolumn{3}{c}{Observed peak (MiB)} \\
\midrule
HumanEval & Native & 4.271 & 4.822 & 17250.1 & 20500.0 & 2009.3 \\
HumanEval & Full CaRR & 4.310 & 4.933 & 17250.4 & 23326.0 & 2058.6 \\
MATH-500 & Native & 3.683 & 4.120 & 17069.4 & 19634.0 & 2009.1 \\
MATH-500 & Full CaRR & 3.798 & 4.316 & 17158.1 & 21248.0 & 2058.6 \\
\bottomrule
\end{tabular}
\caption{Warmed batch-one system measurements on one RTX 5090, BF16, 64 scheduled forwards, using 16 fixed requests per task and three repetitions per arm (192 observations). Timing includes actual final-result JSON serialization and append/flush/close, as well as generation, necessary history, decoding, data movement and original SCR. Loading, warmup, offline scoring and optional trace archival are excluded. CPU RSS is sampled every 10 ms; reserved GPU memory includes allocator caching.}
\label{tab:carr_discussion_system_cost}
\end{table}

\paragraph{Measured system costs.}
Table~\ref{tab:carr_discussion_system_cost} reports request processing through
materialized submission and normal result serialization/write. We use the
submitted old-token gate and original SCR, without the syntax guard, in the
Full arm. Public Code metadata is cached, and no unit tests are executed during
inference. Each arm/repetition runs in a new process, with two excluded warmup
requests. The measured history remains in RAM; optional audit-trace archival is
outside the timing boundary. Stream flush/close does not request durable
\texttt{fsync}. Allocated GPU memory measures live tensors; reserved memory also
includes allocator caching. Their differences include the full pipeline and
sequence-length effects, not solely trajectory storage.

For each request, we take the median latency over three repetitions, form the
Full/Native ratio, and then report the median of these request ratios. This gives
$1.0075$ on HumanEval and $1.0337$ on MATH-500. These are distinct from ratios
of the pooled latency medians. The percentiles in the table describe this fixed
subset and measurement configuration.

Full minus Native differences between the observed peak allocated/reserved/RSS
values are $+0.2/+2826.0/+49.3$ MiB on HumanEval and
$+88.6/+1614.0/+49.5$ MiB on MATH-500.

\endgroup

\begingroup\raggedbottom
\section{Additional Selection Controls}
\label{app:discussion-gates}
We evaluate actual generation from the prompt on all 164 HumanEval and 500 MATH-500 requests at $1/8$: 48 mask and 16 edit forwards, with 17 saved edit states. Nine policies give 5,976 generated trajectories. Each is evaluated as a terminal submission and with the same original SCR; these are two endpoints of the same generated trajectory. The checkpoint, P0 prompts, 512-token initial mask, precision, request order and packed-batch rule are fixed. Prompt, batch and initial-draft identity are checked across policies. These two-benchmark controls report three quality metrics and are separate from the original four-benchmark six-metric average.

All TCR variants retain the original eligibility conditions, lineage handling, control-token exclusions, per-position safeguards and last-step constraint. The original gate uses old-token probability below 0.30. Removing that threshold retains the original cap and cap ranking. Proposed confidence selects low pre-intervention proposed-token probability; entropy selects high pre-intervention entropy; the random control uses seed 0. For each of these ranking controls, $K$ is the number the original gate would select, including its cap, on that policy's current context. Ranking then selects $K$ positions from the same eligible pool. No MASK probability is used.

The joint rankings sum stable ordinal ranks of low old-token probability with low proposed-token confidence or high entropy, using equal weights and the same current-context $K$. The conjunction requires old probability below 0.30 and proposed probability below 0.50; it therefore changes intervention intensity. These definitions were fixed before their generation outputs were observed. Subsequent contexts and rollout intervention totals can differ.

\begin{table}[H]

\centering
\small
\setlength{\tabcolsep}{3pt}
\begin{tabular}{lrrrrrr}
\toprule
& \multicolumn{3}{c}{Terminal submission} & \multicolumn{3}{c}{Original SCR} \\
Generation policy & HE & HE+ & MATH & HE & HE+ & MATH \\
\midrule
Native & 23.17 & 20.73 & 33.80 & 26.22 & 23.78 & 35.00 \\
Original old-token & 25.61 & 23.17 & 34.80 & 31.10 & 28.05 & 36.00 \\
No old threshold & 28.05 & 26.22 & 35.20 & 32.32 & 29.88 & 35.80 \\
Proposed confidence & 28.66 & 25.61 & 35.80 & 31.71 & 28.66 & 37.00 \\
Entropy & 29.27 & 26.83 & 35.20 & 32.93 & 29.88 & 36.20 \\
Random (seed 0) & 25.61 & 23.78 & 35.00 & 29.88 & 28.05 & 36.40 \\
Old + proposed rank & 26.22 & 23.78 & 35.80 & 31.71 & 28.66 & 36.80 \\
Old + entropy rank & 26.22 & 23.78 & 35.60 & 31.71 & 28.66 & 36.60 \\
Old + confidence veto & 25.61 & 23.17 & 35.40 & 29.27 & 26.22 & 36.60 \\
\bottomrule
\end{tabular}
\caption{Actual-generation controls at $1/8$ (64 scheduled forwards). HE/HE+ use 164 HumanEval requests; MATH uses all 500 MATH-500 requests. Values are percentages. HE+ requires the same completion to pass Base and additional tests. Original SCR is identical across rows; Native+SCR is SCR-only and Original old-token+SCR is submitted Full CaRR. Ranking controls match the original gate count on each current context. The confidence-veto conjunction changes intervention intensity.}
\label{tab:carr_discussion_gate_controls}
\end{table}

\begin{table}[H]

\centering
\small
\begin{tabular}{lrr}
\toprule
Generation policy & HumanEval & MATH-500 \\
\midrule
Native & 0.00 & 0.00 \\
Original Full (old-token) & 34.49 & 37.45 \\
No old threshold & 35.62 & 38.35 \\
Proposed confidence & 34.20 & 36.94 \\
Entropy & 34.46 & 36.70 \\
Random (seed 0) & 34.80 & 37.38 \\
Old + proposed rank & 34.10 & 37.25 \\
Old + entropy rank & 34.36 & 37.22 \\
Old + confidence conjunction & 29.37 & 32.60 \\
\bottomrule
\end{tabular}
\caption{Mean remasks per request over the complete rollout. Matching the original gate count on each current context does not force equal rollout totals after trajectories diverge.}
\label{tab:carr_gate_interventions}
\end{table}

The controls support deferred refill across several fixed selection choices. Joint rankings raise the submitted Full point estimates, while proposal-only matches or exceeds them here. The confidence conjunction gives lower HumanEval and higher MATH-500 accuracy with original SCR. Paired descriptive bootstrap intervals for joint-versus-original Full differences include zero; the table is not used to select a new best policy. Per-request submissions, scores, intervention records and paired rescue/break summaries are retained with the experiment configuration.

The canonicalization syntax guard changes one submission in each Math trajectory group and leaves all accuracy entries in Table~\ref{tab:carr_discussion_gate_controls} unchanged. It is a separately reported sensitivity analysis; all primary SCR columns use the original policy.

\section{Math Rule Roles, Simplification and Syntax Correction}
\label{app:discussion-math}
The core operations are deferred replacement refill and trajectory-based submission recovery. Proposal alignment, lineage tracking, control-token exclusion and disabling terminal remasking implement these operations within the finite schedule. Numerical thresholds, caps, the per-token count limit, structural radius, persistence, dominance and edit-distance conditions are empirical settings. The cap and count limit bound intervention; the radius separates remasking from nearby structural edits. Two-state persistence asks for cross-state evidence, strict dominance resolves conflicting answer groups, and distance one or two restricts the correction to a local change. The subset comparison below tests branch removal while retaining these settings; it does not establish the necessity of each numerical value.

The four-branch subset retains same-final, persistent earlier-box, terminal-basin recovery and rational canonicalization. It removes edit-local and answer-linked consensus, preserving every retained predicate and precedence; removal exposes the next eligible rule. Across the audited arms and budgets, full-to-subset accuracy changes are at most 0.61 on GSM8K and 0.20 on MATH-500. The comparison tests branch removal on saved trajectories, including contract-development evidence and repeated questions across budgets.

Edit-local winning interventions give 17 rescues and 14 breaks relative to terminal outputs. Answer-linked consensus is eligible three times and always superseded. These counts describe branch behavior; Table~\ref{tab:carr_math_rule_subset} gives the actual subset comparison.
\begin{table}[H]

\centering
\small
\begin{tabular}{lrrrr}
\toprule
Trajectory / SCR policy & $1/8$ & $1/4$ & $1/2$ & $1/1$ \\
\midrule
Native / Original full & 64.90 / 35.00 & 77.56 / 43.20 & 84.31 / 48.60 & 85.14 / 50.60 \\
Native / Four branches & 64.29 / 34.80 & 77.63 / 43.20 & 84.38 / 48.80 & 85.22 / 50.60 \\
Native / Full + guard & 64.90 / 35.00 & 77.56 / 43.20 & 84.31 / 48.80 & 85.14 / 51.00 \\
Native / Four branches + guard & 64.29 / 34.80 & 77.63 / 43.20 & 84.38 / 49.00 & 85.22 / 51.00 \\
TCR / Original full & 69.75 / 36.00 & 79.91 / 44.00 & 84.91 / 48.60 & 85.82 / 50.20 \\
TCR / Four branches & 69.98 / 35.80 & 79.76 / 44.00 & 84.91 / 48.80 & 85.90 / 50.20 \\
TCR / Full + guard & 69.75 / 36.00 & 79.91 / 44.00 & 84.91 / 48.80 & 85.82 / 50.60 \\
TCR / Four branches + guard & 69.98 / 35.80 & 79.76 / 44.00 & 84.91 / 49.00 & 85.90 / 50.60 \\
\bottomrule
\end{tabular}
\caption{Rule-subset analysis on saved trajectories. Each cell is GSM8K / MATH-500 accuracy (\%), with 1,319 / 500 requests per budget. Four branches retain same-final, persistent earlier-box, terminal-basin recovery and rational canonicalization, preserving retained predicates and precedence. The separate syntax guard declines ambiguous mixed-number canonicalization and continues the remaining frozen branches. The analysis includes contract-development evidence and repeated questions across budgets.}
\label{tab:carr_math_rule_subset}
\end{table}

\paragraph{Boxed-only replay at $1/8$.}
The boxed-only policy preserves the terminal response when its
last boxed occurrence is closed, nonempty, and free of model
control markers. Otherwise, it extracts the last eligible
complete box from the terminal state, then tries the nearest
earlier state containing an eligible box. A recovered submission
contains the selected boxed payload with its text preserved.
If recovery finds no eligible box, the original terminal response
is retained. Table~\ref{tab:carr_boxed_only_replay} compares this
policy with the four-branch subset and original Full SCR on
the same saved trajectories.

\begin{table}[H]

\centering
\small
\caption{Accuracy (\%) on the same saved $1/8$ trajectories.
GSM8K and MATH-500 contain 1,319 and 500 requests per generation
arm, respectively. Four denotes the four-branch subset;
Full denotes original SCR.}
\label{tab:carr_boxed_only_replay}
\begin{tabular}{llrrrr}
\toprule
Task & Arm & Terminal & Boxed-only & Four & Full \\
\midrule
GSM8K    & Native & 63.99 & 63.99 & 64.29 & 64.90 \\
GSM8K    & TCR    & 69.83 & 69.83 & 69.98 & 69.75 \\
MATH-500 & Native & 33.80 & 33.80 & 34.80 & 35.00 \\
MATH-500 & TCR    & 34.80 & 35.00 & 35.80 & 36.00 \\
\bottomrule
\end{tabular}
\end{table}
\paragraph{Canonicalization syntax correction.}
The shared expression parser interprets integer--TeX-fraction juxtaposition,
such as \verb|5\dfrac{1}{2}|, as multiplication. A conservative canonicalization
guard declines that rewrite and continues the remaining frozen branches.
Across the two original generation arms and four budgets, it changes ten
submissions, including repeated questions, correcting six errors and introducing
none under the original evaluator. GSM8K and the $1/8$ and $1/4$ headline results
are unchanged; MATH-500 increases by 0.20 and 0.40 at the
$1/2$ and $1/1$ budgets, respectively, on both arms. Original and guarded results are reported separately in
Table~\ref{tab:carr_math_rule_subset}.

A separate conservative mixed-syntax abstention at the shared-parser entry
points was replayed across all 14,552 historical Math trajectory instances,
with original-policy output parity verified for every instance. It gives the
same ten submission changes, with no additional changes from other consumers
under this candidate. This replay addresses the recorded impact of that syntax
case. On the additional nine gate trajectory groups, the canonicalization guard
changes one submission per Math group and leaves all nine accuracy entries
unchanged; these groups are reported separately from the historical correction.

\endgroup

\end{document}